\documentclass[conference]{IEEEtran}
    \IEEEoverridecommandlockouts
    \usepackage[numbers,sort&compress]{natbib}
    \usepackage{url}
    \usepackage{booktabs}
    \usepackage{array}
    \usepackage{tabularx}
    \usepackage{algorithm}
    \usepackage{algpseudocode}
    \usepackage{graphicx}
    \usepackage{xcolor}
    \usepackage{multirow}
    \usepackage[most]{tcolorbox}
    \usepackage{latexsym}
    \usepackage{microtype}
    \usepackage{inconsolata}
    \usepackage{pifont}
    \usepackage{hyperref}
    \usepackage{orcidlink}
    \def\BibTeX{{\rm B\kern-.05em{\sc i\kern-.025em b}\kern-.08em
        T\kern-.1667em\lower.7ex\hbox{E}\kern-.125emX}}

\begin{document}
    
    \title{Question-to-Knowledge (Q2K): Multi-Agent Generation of Inspectable Facts for Product Mapping\\
    
    }
    
    \author{\IEEEauthorblockN{Wonduk Seo*\thanks{The authors denoted as * have contributed equally to this work.} \orcidlink{0009-0008-6070-1833}}
    \IEEEauthorblockA{AI Research, Enhans \\
    Seoul, South Korea \\
    wonduk@enhans.ai\\
    }
    \and
    \IEEEauthorblockN{Taesub Shin* \orcidlink{0009-0000-5714-592X}}
    \IEEEauthorblockA{AI Research, Enhans \\
    Seoul, South Korea \\
    taesub@enhans.ai}
    \and
    \IEEEauthorblockN{Hyunjin An \orcidlink{0009-0005-5102-9983}}
    \IEEEauthorblockA{AI Research, Enhans \\
    Seoul, South Korea \\
    hyunjin@enhans.ai}
    \and
    \IEEEauthorblockN{Dokyun Kim \orcidlink{0000-0002-5636-2743}}
    \IEEEauthorblockA{AI Research, Enhans \\
    Seoul, South Korea \\
    dokyun@enhans.ai}
    \and
    \IEEEauthorblockN{Seunghyun Lee†\thanks{† denotes corresponding author.} \orcidlink{0009-0000-1687-1597}}
    \IEEEauthorblockA{AI Research, Enhans \\
    Seoul, South Korea \\
    seunghyun@enhans.ai}
    }
    
    \maketitle
    
    \begin{abstract}
    Identifying whether two product listings refer to the same Stock Keeping Unit (SKU) is a persistent challenge in e-commerce, especially when explicit identifiers are missing and product names vary widely across platforms. Rule-based heuristics and keyword similarity often misclassify products by overlooking subtle distinctions in brand, specification, or bundle configuration. To overcome these limitations, we propose \emph{Question-to-Knowledge (Q2K)}, a multi-agent framework that leverages Large Language Models (LLMs) for reliable SKU mapping. Q2K integrates: (1) a \emph{Reasoning Agent} that generates targeted disambiguation questions, (2) a \emph{Knowledge Agent} that resolves them via focused web searches, and (3) a \emph{Deduplication Agent} that reuses validated reasoning traces to reduce redundancy and ensure consistency. A human-in-the-loop mechanism further refines uncertain cases. Experiments on real-world consumer goods datasets show that Q2K surpasses strong baselines, achieving higher accuracy and robustness in difficult scenarios such as bundle identification and brand–origin disambiguation. By reusing retrieved reasoning instead of issuing repeated searches, Q2K balances accuracy with efficiency, offering a scalable and interpretable solution for product integration.\footnote{Code is available at: \url{https://github.com/viralpick/paper-q2k-artifact}.}
    \end{abstract}
    
    \section{Introduction}
    The explosive growth of e-commerce platforms has brought renewed attention to the problem of product mapping, the task of determining whether two product listings correspond to the same Stock Keeping Unit (SKU). Accurate SKU mapping underpins critical applications such as multi-platform price monitoring, sourcing optimization, and bundled product recommendation. However, this task remains highly challenging: (1) product identifiers such as barcodes are often missing or inconsistently used, and (2) product names across platforms vary widely in structure, terminology, and granularity. Even small differences in attributes such as brand, variant, specification, or quantity can change whether two listings refer to identical products. This ambiguity makes reliable SKU mapping one of the most pressing technical bottlenecks in modern e-commerce pipelines~\cite{mackova2023promap,akritidis2018effective,primpeli2019wdc}.
    
    Existing approaches to SKU mapping largely fall into rule-based systems and keyword-driven similarity methods. Rule-based systems rely on explicit heuristics, such as string normalization, brand token matching, or quantity checks. While effective in narrow cases, these methods struggle to generalize across product categories and cannot capture nuanced distinctions~\cite{kopcke2010evaluation}. Keyword similarity methods, on the other hand, often overemphasize surface overlaps while ignoring semantic differences in brand origin, bundle configuration, or specifications. As a result, both approaches frequently yield false positives or negatives, particularly in ambiguous product families and bundled offerings~\cite{ubrangala2024searching}.
    
    Instead of pursuing model fine-tuning as existing studies suggest~\cite{xue2023pumgpt,de2022multi,cheng2024commerce,qiu2021query}, we take a different direction: \emph{Can a new SKU mapping framework be built directly on top of general-purpose LLMs, without costly and inflexible retraining?} Continuous fine-tuning is not only expensive but  hard to maintain under the dynamic nature of e-commerce, where product catalogs evolve rapidly and diverge from training distributions. Moreover, SKU mapping inherently requires access to up-to-date product information, making web search indispensable: attributes such as availability, branding, or origin change frequently, and static models cannot reliably capture these shifts. Our goal is to treat the LLM as a flexible reasoning module that can adapt immediately to new products, remain cost-efficient, and avoid the overhead of retraining with each update.
    
    To address these challenges, we propose \emph{Question-to-Knowledge (Q2K)}, a novel multi-agent framework that transforms SKU mapping into a process of generating and validating inspectable facts. Compared to prior methods that rely solely on surface similarity or opaque model predictions and motivated by recent advances in query reformulation utilizing LLMs~\cite{wang2023query2doc,alaofi2023can,jia2024mill,nachimovsky2025multi,seo2025qa}, Q2K explicitly decomposes the mapping task into three coordinated agents:
    
    \paragraph{Reasoning Agent} which analyzes a base--candidate product pair and generates targeted attribute-specific disambiguation questions (e.g., brand, origin, specification, quantity) to identify missing or ambiguous features;
    
    \paragraph{Knowledge Agent} which conducts focused web searches to resolve these questions by synthesizing authoritative evidence into concise, self-contained answers;
    
    \paragraph{Deduplication Agent} which leverages dense retrieval to detect and reuse previously stored question--answer reasoning traces, reducing redundancy, lowering cost, and ensuring consistency across mappings.
    
    Furthermore, a human-in-the-loop mechanism enhances Q2K by validating low-confidence cases, with corrections fed back into the reasoning database for continuous improvement. This design allows Q2K to produce transparent, reusable, and cost-efficient reasoning chains rather than opaque, one-off judgments.
    
    To assess the effectiveness of Q2K, we evaluate it against $4$ representative baselines:  
    (1) a rule-based mapping system combining string normalization, brand alignment, and quantity matching heuristics;  
    (2) zero-shot inference, where models directly judge equivalence from product names;  
    (3) few-shot inference, which augments prompts with a small set of labeled examples;  
    (4) web-search inference, where models are allowed to query the web without structured decomposition.
    
    Across real-world consumer goods datasets, Q2K consistently outperforms strong baselines, delivering significant improvements in SKU mapping accuracy. Crucially, by retrieving top-$k$ relevant reasoning traces rather than repeatedly issuing new web searches, Q2K lowers computational and operational costs while preserving robustness in challenging cases such as bundle identification, brand--origin disambiguation, and multi-source sourcing. These results highlight Q2K’s ability to tackle one of the most persistent challenges in e-commerce. By reframing SKU mapping as an interpretable, retrieval-augmented reasoning process, Q2K combines automation with human oversight to enable scalable and trustworthy product integration across heterogeneous platforms.

    \section{Data Collection}
    We constructed a dataset of product name pairs through a semi-manual collection process, designed in close collaboration with our industry partner. The collection procedure involved specialized product sellers as human annotators, who possess domain expertise in distinguishing subtle variations across brands, specifications, and bundled offerings. To further ensure data quality, their annotations were regularly cross-validated through supervision checks conducted by senior reviewers.

    \subsection{Collection Procedure}
    The client initially provided a list of reference product names, denoted as \textit{base\_product}. Human data collectors were then instructed to search a designated e-commerce website and gather a list of potentially matching product names. For each candidate, the collector extracted the product title and recorded it as \textit{compared\_product}. 
    
    In addition to recording potential matches, the collectors explicitly flagged products that did not correspond to the given \textit{base\_product}. These flagged items served as negative examples for subsequent product mapping. To further ensure accuracy, a supervisor reviewed the flagged items and validated whether they were labeled correctly.
    
    \subsection{Preprocessing}
    Minimal preprocessing was applied to maintain fidelity to the raw data. Each instance was represented as a pair:
    \[
    (\textit{base\_product}, \textit{compared\_product}).
    \]
    No additional normalization (e.g., tokenization, stopword removal, or lemmatization) was performed at this stage in order to preserve the original wording of product listings.
    
    \subsection{Dataset Statistics}
    The resulting dataset contains approximately $72,250$ rows of product pairs. Notably, the majority of collected samples fall within the food \& beverage (F\&B) domain. Based on manual inspection, we estimate that around $3\%$--$5\%$ of the pairs may contain labeling errors, largely due to subtle attribute differences that were difficult for annotators to resolve.  
    
    \begin{table}[!htbp]
    \centering
    \caption{Example of a mislabeled pair. Despite sharing identical dosage and packaging, the pair was annotated as non-equivalent since the compared product emphasized its Swiss origin. In reality, both refer to the same product line, highlighting the difficulty of annotation.}
    \begin{tabular}{p{0.45\linewidth} p{0.45\linewidth}}
    \toprule
    \textbf{Base Product} & \textbf{Compared Product} \\
    \midrule
    MegaDoseD Vitamin D3 4000IU, 120 tablets × 3 (12 months) &
    Korea Eundan MegaDoseD Vitamin D3 4000IU Swiss-made, 120 tablets × 3 (12 months) \\
    \bottomrule
    \end{tabular}
    \label{tab:annotation_example}
    \end{table}
    
    This example (see Table~\ref{tab:annotation_example}) illustrates the reason why SKU mapping is inherently challenging: optional descriptors such as origin, supplier emphasis, or promotional claims may appear decisive but do not always indicate genuine product differences. Despite such cases, the dataset remains a representative benchmark for evaluating the proposed framework in realistic e-commerce settings.

    \begin{figure*}[!htpb]
        \centering
        \includegraphics[width=1\textwidth]{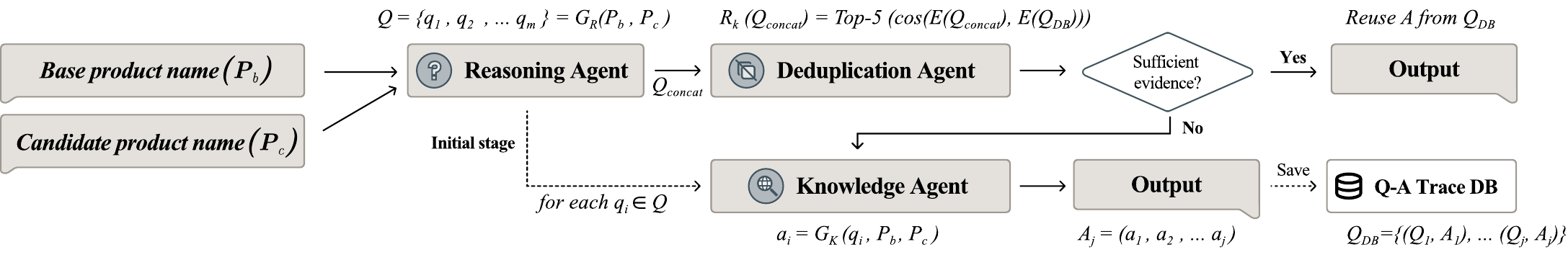}
        \caption{\textbf{Overview of the Q2K workflow}. The Reasoning Agent generates disambiguation questions from a product-pair, the Deduplication Agent retrieves top-k reasoning traces and checks information gain, and the Knowledge Agent provides authoritative answers when evidence is insufficient.}
        \label{fig:main}
    \end{figure*}
    
\section{Methodology}
Our proposed framework \textbf{Q2K} operates through the collaboration of $3$ specialized agents: (1) a \emph{Reasoning Agent}, (2) a \emph{Knowledge Agent}, (3) a \emph{Deduplication Agent}, and followed by a final \emph{decision stage} that integrates all evidence to determine SKU equivalence.\footnote{Prompt templates for all components, including the Reasoning, Knowledge, and Deduplication agents, as well as the final decision stage, are provided in Section~\ref{sec:prompt-design}.} An overview of the framework is illustrated in Figure~\ref{fig:main}.

\subsection{Preliminaries}
We denote each product-pair instance as
\[
(P_b, P_c, y),
\]
where \(P_b\) is the \emph{base} product name, \(P_c\) is the \emph{candidate} (or compared) product name, and \(y \in \{0,1\}\) is the ground-truth label indicating whether the two products correspond to the same SKU (\(y=1\)) or not (\(y=0\)). The full dataset is written as
\[
\mathcal{D} = \{(P_b^{(i)}, P_c^{(i)}, y^{(i)})\}_{i=1}^{N}.
\]

Throughout this section, we use \(G\) to denote LLM-based agents with different subscripts indicating their functional roles. Specifically, \(G_\mathcal{R}\) is the \emph{Reasoning Agent}, which takes a product pair \((P_b, P_c)\) as input and generates a set of disambiguation questions \(\mathcal{Q}\). \(G_\mathcal{K}\) is the \emph{Knowledge Agent}, which answers each question \(q_i \in \mathcal{Q}\) using web evidence and outputs an answer \(a_i\). \(G_\mathcal{D}\) is the \emph{Deduplication Agent}, which decides whether previously stored reasoning traces can be reused for a new product pair, and \(G_\mathcal{F}\) denotes the final \emph{decision module} that aggregates all signals and predicts SKU equivalence. 

We represent the set of questions generated for a given pair as
\(\mathcal{Q} = \{q_1, q_2, \dots, q_m\}\), and the corresponding answer set as
\(\mathcal{A} = \{a_1, a_2, \dots, a_m\}\). To enable reasoning reuse, we maintain a knowledge repository
\[
\mathcal{Q}_{\text{DB}} = \{(Q_1, A_1), \dots, (Q_j, A_j)\},
\]
where each \(Q_i\) is a concatenated question sequence and \(A_i\) is its validated answer set. An embedding function \(E(\cdot)\) maps text into a vector space, and cosine similarity is used to retrieve the top-$k$ most similar reasoning traces, denoted by the operator \(R_k(\cdot)\) in Eq.~\eqref{eq:retrieval}. We focus on five matching dimensions: brand, core product name, variant, specification, and quantity as summarized in Table~\ref{tab:matching-rules}, which guide both question generation and final decision-making.

\subsection{Reasoning Agent}
Product mapping often fails due to missing or ambiguous attribute information, such as unclear manufacturer origin, inconsistent quantity representation, or variant specifications. Motivated by this observation, the \emph{Reasoning Agent} decomposes each product-pair comparison into targeted, attribute-specific questions. Given a base product name \(P_b\) and a candidate product name \(P_c\), the agent generates a set of disambiguation questions:
\begin{equation}\label{eq:reasoning-generation}
\mathcal{Q} = \{q_1, q_2, \dots, q_m\} = G_\mathcal{R}(P_b, P_c),
\end{equation}
where \(G_\mathcal{R}\) denotes the \emph{Reasoning Agent}, and \(m\) is dynamically determined based on the ambiguity detected in the product pair. Each generated question explicitly corresponds to one of five matching dimensions: \emph{brand}, \emph{core product name}, \emph{variant}, \emph{specification}, or \emph{quantity} (see Table~\ref{tab:matching-rules}).

\begin{table}[!htbp]
    \centering
    \caption{Five matching dimensions used by the \emph{Reasoning Agent} to generate targeted disambiguation questions for SKU mapping.}
    \begin{tabular}{lp{0.65\linewidth}}
    \toprule
        \textbf{Attribute} & \textbf{Role in Question Generation} \\
    \midrule
        Brand & Ensures consistency of manufacturer, label, or maker across both products. \\
        Core Product Name & Identifies whether names are synonyms or distinct product lines. \\
        Variant & Clarifies options such as flavor, color, or finish. \\
        Specification & Validates differences in size, weight, volume, or technical details. \\
        Quantity & Confirms pack size, unit count, or bundle configuration. \\
    \bottomrule
    \end{tabular}
    \label{tab:matching-rules}
\end{table}

\subsection{Knowledge Agent}
While the Reasoning Agent formulates targeted questions, the \emph{Knowledge Agent} retrieves factual evidence to resolve them. For each generated question \(q_i \in \mathcal{Q}\), the Knowledge Agent executes focused web queries and synthesizes authoritative answers into structured outputs:
\begin{equation}\label{eq:knowledge-answer}
a_i = G_\mathcal{K}(q_i, P_b, P_c),
\end{equation}
where \(G_\mathcal{K}\) denotes the \emph{Knowledge Agent}, and \(a_i\) is a concise, self-contained explanation oriented toward product equivalence checking. All answers are compiled into an evidence set:
\begin{equation}\label{eq:knowledge-set}
\mathcal{A} = \{a_1, a_2, \dots, a_m\}.
\end{equation}
This process ensures that ambiguous attributes are resolved with external evidence, rather than relying solely on LLM priors. By grounding reasoning in inspectable facts, Q2K supports both automation and human validation.

\subsection{Deduplication Agent}
Naively executing web searches for every product-pair comparison introduces redundancy and high operational cost. To address this, the \emph{Deduplication Agent} coordinates two key operations: (1) retrieving potentially relevant reasoning traces, and (2) deciding whether these traces provide sufficient information to resolve the current case.  

First, instead of treating each question independently, the set of generated questions \(\{q_1, q_2, \dots, q_m\}\) is concatenated into a structured sequence:

\[
Q_{\text{concat}} = \text{``Question1: } q_1\text{;} \dots \text{; Question}m: q_m\text{''}.
\]

This concatenated query string is embedded using a sentence transformer and compared against the repository of previously stored reasoning traces. The top-$k$ most semantically similar entries are retrieved:

\begin{equation}\label{eq:retrieval}
R_k(Q_{\text{concat}}) = \operatorname{Top-}k \left( \cos(E(Q_{\text{concat}}), E(\mathcal{Q}_{\text{DB}})) \right),
\end{equation}

where \(E(\cdot)\) is the embedding function and \(\mathcal{Q}_{\text{DB}}\) denotes the repository of concatenated question--answer reasoning traces. Formally, we define
\[
\mathcal{Q}_{\text{DB}} = \{(Q_1, A_1), (Q_2, A_2), \dots, (Q_j, A_j)\},
\]
where each \(Q_i\) is a concatenated sequence of disambiguation questions and \(A_i\) is its corresponding validated answer set. Thus, \(\mathcal{Q}_{\text{DB}}\) represents $j$ pre-saved knowledge pairs that can be directly reused to resolve new product-pair comparisons.

The \emph{Deduplication Agent} subsequently evaluates whether the retrieved traces provide sufficient \emph{information gain} to support equivalence reasoning. If the retrieved evidence is adequate, the stored answers are directly reused, ensuring both efficiency and consistency. If the evidence is insufficient, the agent escalates to the full Reasoning--Knowledge process: new disambiguation questions are issued, authoritative answers are synthesized via web search, and the resulting reasoning chain is added back into the knowledge base. 

\subsection{Decision Stage}
After evidence aggregation, Q2K performs a final decision-making stage that integrates all available signals: the base and candidate product descriptions, disambiguation questions, and validated knowledge traces. The model evaluates factual consistency across the five matching dimensions (brand, core product name, variant, specification, and quantity) and outputs a binary decision (\(1 =\) same SKU, \(0 =\) different SKU).  
This stage preserves interpretability through explicit evidence citation. It thereby transforms accumulated reasoning into a transparent and verifiable mapping judgment rather than a black-box classification.

    \section{Experiments}
    \subsection{Setup}
    We conduct experiments on our collected SKU mapping dataset to evaluate the effectiveness of the proposed \emph{Question-to-Knowledge (Q2K)} framework. All experiments are designed to ensure fairness in comparison with baseline methods by using the same product-pair inputs and evaluation protocol.
    
    \subsubsection{LLM and Retrieval Models}
    We use \textbf{GPT-4o}~\cite{achiam2023gpt} for all three agents in Q2K, accessed via API with temperature fixed at $0$ to ensure deterministic and reproducible outputs.
    
    For retrieval, we employ \textit{text-embedding-3-small} model from OpenAI~\cite{achiam2023gpt} to embed concatenated question sequences. The top-$5$ most similar reasoning traces are retrieved from the knowledge repository for ensuring a balance between evidence coverage and efficiency.
    
    \subsubsection{Evaluation Metrics}
    Since the SKU mapping task is framed as a binary classification problem (equivalent vs. non-equivalent), we report \emph{accuracy} as the primary evaluation metric. Formally, given $N$ product pairs with ground-truth labels $y_i \in \{0,1\}$ and predicted labels $\hat{y}_i$, accuracy is defined as:
    \begin{equation}
    \text{Accuracy} = \frac{1}{N} \sum_{i=1}^{N} \mathbf{1}\{y_i = \hat{y}_i\},
    \end{equation}
    where $\mathbf{1}\{\cdot\}$ is the indicator function. Accuracy directly reflects the proportion of correctly mapped product pairs.

\subsubsection{Prompt Design}
\label{sec:prompt-design}
To enable interpretable collaboration across the agents in \textbf{Q2K}, we carefully designed four role-specific prompts that reflect the reasoning flow of SKU mapping: reasoning generation, deknowledge evaluation, factual retrieval, and final decision making.\footnote{For brevity, we include simplified versions. Complete prompt templates are available at: \url{https://github.com/viralpick/paper-q2k-artifact}.}

\begin{enumerate}
    \item \emph{Reasoning Agent Prompt}:  
    Guides the LLM to act as a \emph{Product-Matching Expert}. It receives a base product and a candidate product, identifies which of the five attributes (brand, core name, variant, specification, quantity) are ambiguous, and generates up to three targeted disambiguation questions. The output is a JSON object with a concise \texttt{"thinking"} field describing attribute ambiguity and a \texttt{"questions"} array for structured follow-up queries. This prompt ensures controlled reasoning decomposition at the earliest stage of Q2K.
    
    \item \emph{Deduplication Agent Prompt}:  
    Functions as a \emph{Knowledge Sufficiency Evaluator} that decides whether retrieved reasoning traces from the knowledge database are reusable for new questions. It compares newly generated disambiguation questions with prior question–answer pairs and produces two structured lists: \texttt{"reusable\_answers"} and \texttt{"questions\_needing\_fresh\_search"}. This conservative evaluation enforces factual consistency and minimizes redundant searches.
    
    \item \emph{Knowledge Agent Prompt}:  
    Acts as an \emph{Expert Knowledge Provider}. Upon receiving the current task, input pair, and question, it conducts focused web searches to synthesize concise, evidence-grounded explanations. All claims are cited with standardized reference markers to preserve traceability. The output is a self-contained paragraph suitable for downstream decision reasoning. This prompt ensures factual grounding and verifiable transparency.
    
    \item \emph{Decision Stage Prompt}:  
    Operates as a \emph{Product Matching Decision Maker} that integrates all available evidence, including base and candidate descriptions, disambiguation questions, and retrieved knowledge, to output a final binary prediction (\texttt{1 = same SKU}, \texttt{0 = different SKU}) and a concise explanation under the matching rule. This prompt enforces strict adherence to SKU equivalence criteria, emphasizing clarity, brand integrity, and measurable attribute differences.
\end{enumerate}

    \subsubsection{Baselines}
    We compare Q2K against $4$ baselines:  
    
    \begin{itemize}
        \item \textbf{Rule-Based Matching}:  
        This baseline implements a deterministic pipeline built on handcrafted heuristics commonly used in e-commerce product matching. The procedure consists of three sequential stages:  
        \begin{enumerate}
            \item \emph{String Normalization}: Product titles are lowercased and stripped of punctuation, stopwords, and special characters. Common units and symbols are standardized to canonical forms.  
            \item \emph{Brand Token Matching}: A curated dictionary of brand names is used to detect and align brand tokens in both the base and compared product names. If brand tokens are mismatched or absent, the pair is labeled as non-equivalent.  
            \item \emph{Quantity and Specification Alignment}: Numerical tokens are extracted using regular expressions and compared across products. Bundle-related attributes must align for equivalence.  
        \end{enumerate}
        The final decision rule is: two products are considered equivalent only if their normalized core tokens, detected brand, and extracted numerical specifications all match.
    
        \item \textbf{Zero-Shot Inference}: GPT-4o directly predicts product equivalence from raw product names without additional context. This tests the raw reasoning ability of LLMs. 
        \item \textbf{Few-Shot Inference}: GPT-4o is prompted with a small set of labeled examples before inference. This aims to capture attribute diversity through in-context learning. 
        \item \textbf{Web-Search Inference}: GPT-4o is equipped with unrestricted web search for each product pair, without structured decomposition. This allows factual grounding but incurs higher latency and cost.
    \end{itemize}
    
    \subsection{Experiment Results}

We compare our proposed \emph{Question-to-Knowledge (Q2K)} framework with four baseline methods: (1) \emph{Rule-Based Matching}, (2) \emph{Zero-Shot Inference}, (3) \emph{Few-Shot Inference}, and (4) \emph{Web-Search Inference}. The results, summarized in Table~\ref{tab:exp_results}, reveal consistent trends. Traditional rule-based systems achieve only $67.37\%$ accuracy, reflecting their inability to generalize across diverse product structures. Their reliance on string normalization and handcrafted heuristics often leads to misclassifications when brands are aliased, specifications are abbreviated, or promotional terms obscure key details. In contrast, zero-shot GPT-4o inference markedly improves performance to $90.34\%$, demonstrating that LLMs can capture semantic correspondences beyond surface similarity. However, it still falters in ambiguous cases where brand origin, bundle size, or variant type require factual grounding that cannot be inferred from text alone.

Few-shot prompting further enhances adaptability, achieving $92.68\%$ accuracy by incorporating labeled exemplars into contextual reasoning, but its performance remains constrained by limited sample diversity. The web-search baseline performs better ($93.44\%$) by incorporating real-time evidence, reducing hallucinations and increasing factual correctness. Nonetheless, its unstructured query process introduces redundant searches and inconsistent reasoning, which inflate latency and cost. These results collectively highlight a persistent trade-off: rule-based and LLM-only methods are efficient but shallow, while web-grounded systems are accurate but computationally expensive.

By contrast, \emph{Q2K} achieves $95.62\%$ accuracy, surpassing all baselines while maintaining interpretability and efficiency. Its multi-agent design decomposes SKU mapping into distinct reasoning stages, where the \emph{Reasoning Agent} formulates attribute-specific questions, the \emph{Knowledge Agent} retrieves verifiable evidence, and the \emph{Deduplication Agent} reuses validated reasoning traces across product families. This structure minimizes redundancy\footnote{Detailed formulation of the Deduplication Agent and retrieval process is provided in Section~\ref{sec:deduplication}.} yielding both cost and latency reductions without sacrificing precision. These findings confirm that structured, retrieval-augmented reasoning can outperform traditional similarity-based or fully generative baselines, establishing Q2K as a scalable and interpretable paradigm for reliable SKU mapping.
    
    \begin{table}[!htbp]
    \centering
    \caption{Comparison of SKU mapping accuracy across baseline methods and our proposed framework. Q2K (Ours) achieves the highest accuracy by combining query decomposition with web-based evidence retrieval.}
    \begin{tabular}{l c}
    \toprule
    \textbf{Method} & \textbf{Accuracy (Binary Classification)} \\
    \midrule
    Rule-Based Matching & 0.6737 \\
    Zero-Shot Inference & 0.9034 \\
    Few-Shot Inference & 0.9268 \\
    Web Search Inference & 0.9344 \\
    \textbf{Question-to-Knowledge (Ours)} & \textbf{0.9562} \\
    \bottomrule
    \end{tabular}
    \label{tab:exp_results}
    \end{table}

\subsection{Ablation Study}
To better understand the contribution of each component in \textbf{Q2K}, we conduct an ablation study examining how individual agents and their interactions affect performance and efficiency (see Table~\ref{tab:ablation}). The ultimate goal is to reveal how query decomposition, reasoning reuse, and web-based evidence jointly contribute to the final mapping accuracy. 

\paragraph{Question Generation Behavior.} 
On average, the \emph{Reasoning Agent} generated $1.4$ attribute-specific disambiguation questions per product-pair comparison. This indicates that most pairs contained only one uncertain attribute, typically involving brand aliasing or quantity mismatch. While a minority ($<20\%$) involved multiple ambiguities such as brand–variant or specification–quantity combinations. The small average number of generated questions demonstrates that the agent effectively focuses on the most informative attributes rather than over-decomposing the task, which helps reduce prompt length and overall computation.

\paragraph{Deduplication Efficiency.} 
\label{sec:deduplication}
The \emph{Deduplication Agent} played a key role in reducing redundant computation by identifying when previously validated reasoning traces could be reused. Retrieval was triggered in approximately $22\%$ of total comparisons, meaning that nearly one in five product pairs could be resolved without issuing new web queries. These reused reasoning traces typically came from product families with recurring patterns, where attribute equivalence rules were consistent across multiple listings. In such cases, the system bypassed repeated searches and relied on stored reasoning chains, achieving substantial latency and cost reductions while maintaining decision consistency.

\begin{table}[!htbp]
\centering
\caption{Ablation results of \emph{Q2K} components. The Reasoning Agent focuses on concise decomposition, while the Deduplication Agent reuses prior reasoning in $22\%$ of cases, reducing cost and latency.}
\begin{tabular}{l c}
\toprule
\textbf{Ablation Metric} & \textbf{Value} \\
\midrule
Average questions generated per product pair & $1.4$ \\
Cases requiring multiple questions & Minority (\,$<20\%$\,) \\
Deduplication retrieval activation rate & $22\%$ \\
\bottomrule
\end{tabular}
\label{tab:ablation}
\end{table}

Overall, the ablation analysis shows that the synergy between the Reasoning and Deduplication Agents yields the strongest gains in both interpretability and computational efficiency. By minimizing redundant searches and targeting only ambiguous attributes, Q2K maintains high accuracy while reducing average token and latency costs compared to non-deduplicated reasoning.

\section{Conclusion}
In this paper, we introduced \emph{Question-to-Knowledge (Q2K)}, a multi-agent framework that redefines SKU mapping as a process of structured reasoning rather than surface-level text matching. By decomposing each comparison into targeted questions, retrieving web-based evidence, and reusing verified reasoning traces, Q2K transforms an opaque classification problem into an interpretable and data-efficient reasoning pipeline. Experiments on large-scale consumer goods datasets demonstrate that Q2K achieves $95.62\%$ accuracy, surpassing baselines such as rule-based, zero-shot, few-shot, and web-search baselines, while substantially reducing redundant computation through selective retrieval. 

Beyond accuracy, Q2K offers practical benefits for real-world e-commerce systems: its modular architecture enables continuous adaptation to evolving catalogs, its reasoning database facilitates knowledge reuse, and its human-in-the-loop design ensures traceable decision-making. Together, these characteristics make Q2K not only a competitive SKU mapping framework but also a broader paradigm for interpretable, retrieval-augmented reasoning in product intelligence. 

\section{Limitations and Future Work}
While Q2K demonstrates strong performance and transparency, several limitations remain. First, the current framework depends on the reliability of web sources for factual grounding; noisy or outdated product information may still introduce occasional errors in reasoning. Second, our dataset, though extensive, is concentrated in the food and beverage domain, and future evaluations should expand to diverse product categories such as electronics or apparel to test cross-domain generalization. Third, Although the human-in-the-loop validation is crucial for quality assurance, it introduces latency that may limit large-scale deployment in fully automated environments. 

In future work, we plan to explore (1) integrating trusted product knowledge graphs to reduce dependency on external web searches, and (2) extending Q2K to multilingual and cross-market scenarios where brand aliases and localized variants further complicate SKU equivalence. 

\section*{Acknowledgment}
The authors would like to express their sincere gratitude to the \textbf{Data Pipeline and Quality Assurance (DP\&QA) team} for their dedicated efforts in data collection and validation checks, which were essential to this research. We also thank the \textbf{Enhans Agent team} for their continuous technical support and insightful feedback throughout the project. This research was supported in part by internal research funding from Enhans.

\section*{Author Contributions}
\noindent
\textbf{Wonduk Seo (Co-First Author):} Conceptualization, Study Design, Methodology Development, Formal Analysis, Investigation, Software Implementation, Writing-Original Draft, Writing-Review \& Editing, Project Administration, and Manuscript Preparation.\\
\textbf{Taesub Shin (Co-First Author):} Conceptualization, Study Design, Methodology Development, Formal Analysis, Investigation, and Software Implementation.\\
\textbf{Hyunjin An (Second Author):} Conceptualization, Study Design, Methodology Development, Visualization, Writing-Review \& Editing, and Manuscript Preparation.\\
\textbf{Dokyun Kim (Third Author):} Conceptualization, Study Design, Methodology Development, and Writing-Review \& Editing.\\
\textbf{Seunghyun Lee (Corresponding Author):} Conceptualization, Study Design, Supervision, Project Administration, and Writing-Review \& Editing.\\[4pt]
All authors participated in result validation, discussion of findings, and the final approval of the submitted manuscript.

\section*{Data Availability}
\noindent
The code and sample datasets used in this study are publicly available at: 
\url{https://github.com/viralpick/paper-q2k-artifact}.\\
All resources were utilized in compliance with open-source licenses and ethical data handling standards.

    \bibliographystyle{ieeetr}
    \bibliography{references}
    
    \end{document}